%% file: vaguecorr.tex
\documentclass[12pt]{article}
\usepackage{chicagor}
\usepackage{times}


\input{defn}

\newcommand{\intension}[1]{[\![ #1 ]\!]}
\begin{document}
\newcommand{\newcomment}{\marginpar{**}}

\title{Intransitivity and Vagueness}
\author{Joseph Y. Halpern%
\thanks{Work supported in part by NSF under grant 
CTC-0208535, by ONR under grants  N00014-00-1-03-41 and
N00014-01-10-511, by the DoD Multidisciplinary University Research
Initiative (MURI) program administered by the ONR under
grant N00014-01-1-0795, and by AFOSR under grant F49620-02-1-0101.
A preliminary version of this paper appears in {\em Principles of
Knowledge Representation and Reasoning: Proceedings of the Ninth
International Conference (KR 2004)}.}
\\
   Cornell University\\
   Computer Science Department\\
   Ithaca, NY 14853\\
   halpern@cs.cornell.edu\\
   http://www.cs.cornell.edu/home/halpern}
\maketitle

\begin{abstract}
There are many examples in the literature that suggest that
indistinguishability is intransitive, despite the fact that the
indistinguishability relation is typically taken to be an equivalence
relation (and thus transitive).  It is shown that if 
the uncertainty perception and the question of 
when an agent {\em reports\/} that two things are indistinguishable are both
carefully modeled, the problems disappear, and indistinguishability can
indeed be taken to be an equivalence relation.  Moreover, this model 
also suggests a logic of {\em vagueness\/} that seems to solve many of the
problems related to vagueness discussed in the philosophical literature.
In particular, it is shown here how the logic can  handle
the {\em sorites paradox}.
\end{abstract}

\section{Introduction}
While it seems that indistinguishability should be an equivalence
relation and thus, in particular, transitive, there are many examples in
the literature that suggest otherwise.  For example, tasters cannot
distinguish a cup of coffee with one grain of sugar from one without
sugar, nor, more generally, a cup with $n+1$ grains
of sugar from one with $n$ grains of sugar.  But they can certainly
distinguish a cup with 1,000 grains of sugar from one with no sugar at
all.  

These intransitivities in indistinguishability lead to intransitivities
in preference.  For example, consider someone who prefers coffee with
a teaspoon of sugar to one with no sugar.  Since she cannot distinguish
a cup with $n$ grains from a cup with $n+1$ grains, she is clearly
indifferent between them.  Yet, if a teaspoon of sugar is 1,000 grains,
then she clearly prefers a cup with 1,000 grains to a cup with no sugar.

There is a strong intuition that the indistinguishability relation
should be transitive, as should the relation of equivalence on
preferences.  Indeed, transitivity is implicit in our use of the word
``equivalence'' to describe the relation on preferences.  Moreover, it
is this intuition that forms the basis of the partitional model for
knowledge used in game theory (see, e.g., \cite{Au}) and in the distributed
systems community \cite{FHMV}.  
On the other hand, besides the obvious experimental observations, there
have been arguments going back to at least Poincar{\'e}
\citeyear{Poincare02} that the physical world is not transitive in this
sense. 
In this paper, I try to reconcile our
intuitions about indistinguishability with the experimental
observations, in a way that seems (at least to me) both intuitively
appealing and psychologically plausible.  I then go on to apply the
ideas developed to the problem of {\em vagueness}.  

To understand the vagueness problem, consider the 
well-known {\em sorites paradox\/}:  If $n+1$ grains of sand make a
heap, then so do $n$.  But 1,000,000 grains of sand are clearly a heap,
and 1 grain of sand does not constitute a heap.  
Let {\bf Heap} to be a predicate such that
$\textbf{Heap}(n)$ holds if $n$ grains of sand make a heap.   What is the
extension of {\bf Heap}?  That is, for what subset of natural numbers
does {\bf Heap} hold?  Is this even well defined?  Clearly the set of
numbers for which {\bf Heap} holds is upward closed: if $n$ grains of
sand is a heap, then surely $n+1$ grains of sand is a 
heap. Similarly, the set of grains of sand which are not a heap
is downward closed: if $n$ grains of sand is not a heap, then $n-1$
grains of sand is not a heap.  However, there is a fuzzy middle ground,
which is in part the reason for the paradox.  The relationship of the
vagueness of {\bf Heap} to indistinguishability should be clear: $n$
grains of sand are indistinguishable from $n+1$ grains.  Indeed, just as

{\bf Heap} is a vague predicate, so is the predicate {\bf Sweet}, 
where  {\bf Sweet}($n$) holds if a cup of coffee with $n$ grains of
sugar is sweet.  
So it is not surprising that an approach to dealing with intransitivity
has something to say about vagueness.  

The rest of this paper is organized as follows.  In
Section~\ref{sec:intransitivity} I discuss my solution to the
intransitivity problem.  In Section~\ref{sec:vagueness}, I show how
how this solution can be applied to the problem of vagueness.
There is a huge literature on the vagueness problem.  Perhaps the
best-known approach in the AI literature involves fuzzy logic, but fuzzy
logic represents only a small part of the picture; the number of recent
book-length treatments, including 
\cite{Keefe00,KS96,Sorenson01,Williamson94}, give a sense of the
activity in the area.  I formalize the intuitions discussed in
Section~\ref{sec:intransitivity} using a logic for reasoning about
vague propositions, provide a sound a complete axiomatization for the
logic, and show  how it can deal with problems like the sorites paradox.
I compare my approach to vagueness to some of the
leading alternatives in Section~\ref{sec:other}.  Finally, I conclude
with some discussion in Section~\ref{sec:discussion}.

\section{Intransitivity}\label{sec:intransitivity}
Clearly part of the explanation for the apparent intransitivity in the
sugar example involves differences that are too small to be detected.
But this can't be the whole story.  To understand the issues, imagine a
robot with a simple sensor for sweetness.  The robot ``drinks'' a cup of
coffee and measures 
how sweet it is.  Further imagine that the robot's sensor is 
sensitive only at the 10-grain level.  Formally, this means that a
cup with 0--9 grains results in a sensor reading of 0, 10--19 grains
results in a sensor reading of 1, and so on.  If the situation were
indeed that simple, then indistinguishability would in fact  be an
equivalence relation.  All cups of coffee with 0--9 grains of sugar would
be indistinguishable,  as would cups of coffee with 10--19 grains, and
so on.  However, in this simple setting, a cup of coffee with 
9 grains of sugar would be distinguishable from cups with 10 grains. 

To recover intransitivity requires two more steps.  The first involves
dropping the assumption that the number of grains of sugar uniquely
determines the reading of the sensor.  There are many reasons to drop
this assumption.  For one thing, the robot's sensor may not be
completely reliable; for example, 12 grains of sugar may occasionally
lead to a reading of 0; 8 grains may lead to a reading of 1.  A second
reason is 
that the reading may depend in part on the robot's state.  After drinking
three cups of sweet coffee, the robot's perception of sweetness may be
dulled somewhat, and a cup with 112 grains of sugar may result in a
reading of 10.  A third reason may be due to problems in the robot's
vision system, so that the robot may ``read'' 1 when the sensor actually
says 2.  It is easy to imagine other reasons; the details do not matter
here.  All that matters is what is done about this indeterminacy.
This leads to the second step of my ``solution''.

To simplify the rest of the discussion, assume that the
``indeterminacy'' is less than 4 grains of sugar, so that if there are
actually $n$ grains of sugar, the sensor reading is between
$\lfloor (n-4)/10 \rfloor$ and $\lfloor (n+4)/10 \rfloor$.%
\footnote{$\lfloor x \rfloor$, the floor of $x$, is the largest integer
less than or equal to $x$.  Thus, for example, $\lfloor 3.2 \rfloor = 3$.}
It follows
that two cups of coffee with the same number of grains may result in
readings that are not the same, but they will be at most one apart.  
Moreover, two cups of coffee which differ by one grain of sugar will
also result in readings that differ by at most one.

The robot is asked to compare the sweetness of cups, not sensor
readings.  Thus, we must ask when the robot {\em reports\/} two cups of
coffee as being of equivalent sweetness.  Given the indeterminacy of the
reading, it seems reasonable
that two cups of sugar that result in a sensor reading that differ
by no more than one are reported as indistinguishable, since they could
have come from cups of coffee with the same number of grains of sugar.  
It is immediate that reports of indistinguishability will be
intransitive, even if the sweetness readings themselves clearly
determine an equivalence relation.  Indeed, if the number of grains in
two cups of coffee differs by one, then the two cups will be reported as
equivalent.  But if the number of grains differs by at least eighteen,
then they will be reported as inequivalent.

To sum up, reports of relative sweetness (and, more generally, reports
about perceptions) exhibit intransitivity; it may
well be that there are three cups of sugar such that $a$ and $b$ are
reported as being equivalent in sweetness, as are $b$ and $c$, but $c$
is reported as being sweeter than $a$.  Nevertheless, the 
underlying ``perceived sweetness'' relation can be taken to be
transitive.  However, ``perceived sweetness'' must then
be  taken to be a relation on the taste of a cup of
coffee tried at a particular time, not on the number of grains of sugar
in a cup.  That is, rather than considering a {\bf Sweeter-Than} relation
where {\bf Sweeter-Than}($n,n'$) holds if a cup of coffee with $n$
grains is sweeter than one with $n'$ grains of sugar, we should consider
a {\bf Sweeter-Than$'$} relation, where {\bf
Sweeter-Than$'$}($(c,s),(c',s')$) holds if cup of coffee $c$ tried by
the agent in (subjective) state $s$ (where the state includes the time,
and other 
features of the agent's state, such as how many cups of coffee she has
had recently) is perceived as sweeter than cup of coffee $c'$ tried by
the agent in state $s'$.  The former relation may not be transitive; the
latter is.  But note that the latter relation does not completely
determine when the agent {\em reports\/} $c$ as being sweeter than $c'$.
Intransitivity in reports of perceptions does not necessarily imply
intransitivity in actual perceptions.

\section{Vagueness}\label{sec:vagueness}

The term ``vagueness'' has been used somewhat vaguely in the
literature.  Roughly speaking, a term is said to be vague if its use
varies both between and within speakers.  (According to Williamson
\citeyear{Williamson94}, this interpretation of vagueness goes back at
least to Peirce \citeyear{Peirce31}, and was also used by Black
\citeyear{Black37} and Hempel \citeyear{Hempel39}.)  In the language of
the previous section, $P$ is vague if, for some $a$, some agents may
report $P(a)$ while others may report $\neg P(a)$ and, indeed, the same
agent may sometimes report $P(a)$ and sometimes $\neg P(a)$. 

Vagueness has been applied to what seem to me to be two distinct, but
related, phenomena.  For one thing, it has been applied to predicates
like {\bf Red}, where the different reports may be
attributed in part to there not being an objective notion of what counts
as red.  That is, two agents looking at the same object (under the
same lighting conditions) may disagree as to whether an object is red, 
although typically they will agree.  Vagueness is also applied to 
situations with epistemic uncertainty, as in the case of
a predicate {\bf Crowd}, where {\bf Crowd}($n$) holds if there are at
least $n$ people in a stadium at a particular time.%
\footnote{Of course, there may still be objective uncertainty as to how
to do the count.  For example, does a pregnant woman count as one or
two?  If the answer is ``one'', then if she goes into labor, at what point
does the answer become ``two''?  The point is that even if we assume
that all these details have been worked out, so that there would be be
complete agreement among all agents as to how many people are in the
stadium if they had all the relevant information, there will still in
general be uncertainty as to how many people are in the stadium.  This
uncertainty leads to vagueness.}  
Here there
may be different responses because agents have trouble estimating the
size of a crowd.  I present a model  that distinguishes 
these two sources of vagueness.    Because vagueness is rather slippery,
I also present a formal logic of vagueness. 

\subsection{A Modal Logic of Vagueness: Syntax and Semantics}
To reason about vagueness, I consider a modal logic ${\cal L}^{{\it DR}}_n$
with two families of
modal operators: $R_1, \ldots, R_n$, where $R_i \phi$ is
interpreted as ``agent $i$ reports $\phi$'', and $D_1, \ldots, D_n$,
where $D_i \phi$ is interpreted as ``according to agent
$i$, $\phi$ is definitely the case''.
For simplicity, I consider only a propositional logic; 
there are no difficulties extending to the first-order case.  As the
notation makes clear, I allow multiple agents, since some issues
regarding vagueness (in particular, the fact that different agents may
interpret a vague predicate differently) are best considered in a
multi-agent setting.

Start with a (possibly infinite) set of primitive
propositions. 
More complicated formulas are formed by
closing off under conjunction, negation, and the  modal operators $R_1,
\ldots, R_n$ and $D_1, \ldots, D_n$.  

A {\em vagueness structure\/} $M$ has the form $(W, P_1, \ldots, P_n,
\pi_1, \ldots, 
\pi_n)$, where $P_i$ is a nonempty subset of $W$ for $i = 1, \ldots, n$,
and $\pi_i$ is an interpretation, which associates with each primitive
proposition a subset of $W$.  
Intuitively, $P_i$ consists of the worlds that agent $i$ 
initially considers plausible.
{F}or those used to thinking
probabilistically, the worlds in $P_i$ can be thought of as those that have 
prior probability greater than $\epsilon$ according to agent $i$, for
some fixed $\epsilon \ge 0$.%
\footnote{In general, the worlds that an agent considers plausible
depends on the agent's subjective state.  That is why I have been
careful here to say that $P_i$ consists of the worlds that agent $i$
{\em initially\/} considers plausible.  
$P_i$ should be thought of as modeling the agent $i$'s prior beliefs, before
learning whatever information led to the agent $i$ to its actual subjective
state. 
It should shortly become clear
how the model takes into account the fact that the agent's set of
plausible worlds changes according to the agent's subjective state.}
A simple class of models is obtained by taking $P_i = W$ for $i = 1,
\ldots, n$; however, as we shall see, in the case of multiple agents,
there are advantages to allowing $P_i \ne W$.  Turning to the
truth assignments $\pi_i$, note that 
it is somewhat nonstandard in
modal logic to have a different truth assignment for each agent; this
different truth assignment is intended to capture the intuition that the
truth of formulas like {\bf Sweet} is, to some extent, dependent on the
agent, and not just on objective features of the world.

I assume that $W \subseteq O \times S_1 \times \ldots S_n$, where 
$O$ is a set of objective states, and $S_i$ is a set of subjective
states for agent $i$.  Thus, worlds have the form $(o,s_1, \ldots, s_n)$.
Agent $i$'s subjective state $s_i$ represents $i$'s perception of the world
and everything else about the agent's makeup that  determines the
agent's report.   For example, in the case of the robot with a sensor,
$o$ could be the actual number of grains of sugar in a cup of coffee and
$s_i$ could be the 
reading on the robot's sensor.  Similarly, if the formula in question was
\textbf{Thin(TW)} (``Tim Williamson is thin'', a formula often considered
in \cite{Williamson94}), then $o$ could represent the actual dimensions
of TW, and $s_i$ could represent the agent's perceptions.  Note that $s_i$
could also include information about other features of the situation,
such as the relevant reference group.  (Notions of thinness are clearly
somewhat culture dependent and change over time; what counts as thin
might be very different if TW is a sumo wrestler.)  In addition, $s_i$
could include the agent's cutoff points for deciding what counts as
thin, or what counts as red.  In the case of the robot discussed in
Section~\ref{sec:intransitivity}, the subjective state could include its
rule for deciding when to report something as sweet.%
\footnote{This partition of the world into objective state and
subjective states is based on the ``runs and systems'' framework
introduced in \cite{HFfull} (see \cite{FHMV} for motivation and
discussion).  The framework has been used to analyze problems ranging
from distributed computing \cite{FHMV} to game theory \cite{Hal15} to belief
revision \cite{FrH1}.  More recently, it has been applied to the
Sleeping Beauty problem \cite{Hal35}.}  

If $p$ is a primitive proposition then,
intuitively, 
$(o,s_1, \ldots, s_n) \in \pi_i(p)$ if $i$  would 
consider $p$ true if $i$ knew exactly what the objective situation was
(i.e., if $i$ knew $o$), given
$i$'s possibly subjective judgment of what counts as ``$p$-ness''.
Given this intuition, it should be clear that all that should
matter in 
this evaluation is the objective part of the world, $o$, and (possibly)
agent $i$'s subjective state, $s_i$.   In the case of the robot, whether
$(o,s_1, \ldots, s_n) \in \pi_i(\textbf{Sweet})$ clearly depends on how
many grains of sugar are in the cup of coffee, and may also depend on
the robot's perception of sweetness and its cutoff points for sweetness,
but does not depend on other robots' perceptions of sweetness.  Note
that the robot may give different answers in two different subjective states,
even if the objective state is the same and the robot knows the objective
state, since both its perceptions of sweetness and its cutoff point for
sweetness may be different  in the two subjective states.

I write $w \sim_i w'$ if 
$w$ and $w'$ agree on agent $i$'s subjective state, and I write $w
\sim_o w'$ if $w$ and $w'$ agree on the objective part of the state.
Intuitively, the $\sim_i$ relation can be viewed as describing the
worlds that agent $i$ considers possible.  Put another way, if $w \sim_i
w'$, then $i$ cannot distinguish $w$ from $w'$, given his current information.
Note that the indistinguishability relation is transitive (indeed, it is
an equivalence relation), in keeping
with the discussion in Section~\ref{sec:intransitivity}.
I assume that $\pi_i$ depends only on the objective part of the state
and $i$'s subjective state, so that if $w \in \pi_i(p)$ for a
primitive proposition $p$, and $w \sim_i w'$ and $w \sim_o w'$, then
$w' \in \pi_i(p)$.  
Note that $j$'s state (for $j \ne i$) has no effect on $i$'s determination
of the truth of $p$.  
There may be some primitive propositions whose truth depends only on
the objective part of the state (for example, {\bf Crowd}($n$) is such a
proposition).  If $p$ is such an objective proposition, then 
$\pi_i(p) = \pi_j(p)$ for all agents
$i$ and $j$, and, if $w \sim_o w'$, then $w \in \pi_i(p)$ iff $w' \in
\pi_i(p)$. 

I next define what it means for a formula to be true.  The truth
of formulas is relative to both the agent and the world.  
I write $(M,w,i) \sat \phi$ if $\phi$ is true according to agent $i$ in
world $w$.  In the case of a primitive proposition $p$, 
$$(M,w,i) \sat p \mbox{ \ \ iff \ \ } w \in \pi_i(p).$$
I define $\sat$ for other formulas by induction.  For conjunction and
negation, the definitions are standard:
$$\begin{array}{l}
(M,w,i) \sat \neg \phi \mbox{ \ \ iff \ \ } (M,w,i) \not\sat \phi;\\
(M,w,i) \sat \phi \land \psi \mbox{ \ \ iff \ \ } (M,w,i) \sat\phi \mbox{
and } (M,w,i) \sat \psi.
\end{array}$$

In the semantics for negation, 
I have implicitly assumed that, given the objective situation and
agent $i$'s subjective state, agent $i$ is prepared to say, for every 
primitive proposition $p$, whether or not $p$ holds.  Thus, if $w \notin
\pi_i(p)$, so that agent $i$ would not consider $p$ true given $i$'s
subjective state in $w$ if $i$ knew the objective situation at $w$, then
I am assuming that $i$ would consider $\neg p$ true in this world.
This assumption is
being made mainly for ease of exposition.  It would be easy to modify
the approach to allow agent $i$ to say (given the objective state and
$i$'s subjective state), either ``$p$ holds'', ``$p$ does not hold'',
or ``I am not prepared to say whether $p$ holds or $p$ does not hold''.%
\footnote{The resulting logic would still be two-valued; the primitive
proposition $p$ would be replaced by a family of three
primitive propositions, $p_y$, $p_n$, and $p_{?}$, corresponding to
``$p$ holds'', ``$p$ does not hold'', and ``I am not prepared to say
whether $p$ holds or does not hold'', with a semantic requirement
(which becomes an axiom in the complete axiomatization) stipulating that
exactly one proposition in each such family holds at each world.}
However, what I am explicitly avoiding here is taking a fuzzy-logic like
approach of saying something like ``$p$ is true to degree .3''.  While
the notion of degree of truth is certainly intuitively appealing, it has
other problems.  The most obvious in this context is where the .3 is
coming from.  Even if $p$ is vague, the notion ``$p$ is true to degree
.3'' is precise.  It is not clear that introducing a continuum of
precise propositions to replace the vague proposition $p$ really solves
the problem of vagueness.  Having said that, there is a natural
connection between the approach I am about to present and fuzzy logic;
see Section~\ref{sec:fuzzy}.

Next, I consider the semantics for the modal operators $R_j$, $j = 1,
\ldots, n$.
Recall that $R_j \phi$ is interpreted as ``agent $j$ reports $\phi$.  
Formally, 
I take $R_j \phi$ to be true 
if $\phi$ is true at all plausible states $j$ considers
possible.  Thus,
taking $\R_j(\phi) = \{w': w \sim_j w'\}$, 
$$\begin{array}{l}
(M,w,i) \sat R_j \phi \mbox{ \ \ iff \ \ }
\mbox{\ \ \ } (M,w',j) \sat \phi \mbox{\
for all $w' \in \R_j(w') \inter P_j$}.
\end{array}$$
Of course, for a particular formula $\phi$, an agent may neither report
$\phi$ nor $\neg \phi$.  An agent may not be willing to say either that
TW is thin or that TW is not thin.  
Note that, effectively, the set of plausible states according to agent
$j$ given the agent's subjective state in world $w$ can be viewed as the
worlds in in $P_j$ that are indistinguishable to agent $j$ from $w$.
Essentially, the agent $j$ is updating the worlds that she initially
considers plausible by intersecting them with the worlds she considers
possible, given her subjective state at world $w$.
If $P_j = W$ for all agents $j = 1, \ldots, n$, then it is impossible
for agents to give conflicting reports; that is, the formula $R_i \phi
\land \neg R_j \phi$ would be inconsistent.  By considering only the
plausible worlds when giving the semantics for $R_j$, it is consistent
to have conflicting reports.

Finally, $\phi$ is definitely true at state $w$ if the truth of $\phi$
is determined by the objective state at $w$:
$$\begin{array}{l}
(M,w,i) \sat D_j \phi \mbox{ \ \ iff \ \ }
\mbox{\ \ \ }(M,w',j) \sat \phi \mbox{\
for all $w'$ such that $w \sim_o w'$}.
\end{array}
$$

A formula is said to be {\em agent-independent\/} if its truth is
independent of the agent.  That is, $\phi$ is agent-independent if, for
all 
worlds $w$, $$(M,w,i) \sat \phi \mbox{ \ \ iff \ \ } (M,w,j) \sat \phi.$$
As we observed earlier, 
objective primitive propositions (whose truth depends only on the
objective part of a world) are
agent-independent; it is easy to see that  
formulas of the form $D_j \phi$ and $R_j \phi$ are as well.  
If $\phi$ is agent-independent, then I often write $(M,w) \sat \phi$
rather than $(M,w,i) \sat \phi$.

\subsection{A Modal Logic of Vagueness: Axiomatization and Complexity}
It is easy to see that $R_j$ satisfies the axioms and rules of the modal
logic KD45.%
\footnote{For modal logicians, perhaps the easiest way to see this is to
observe a relation $\R_j$
on worlds can be defined 
consisting of all pairs $(w,w')$ such that $w \sim_j w'$ and
$w' \in P_j$.  This relation, which characterizes the modal operator
$R_j$, is easily seen to be Euclidean and transitive, and thus
determines a modal operator satisfying the axioms of KD45.}
It is also easy to see that $D_j$ satisfies the axioms of KD45.  
It would seem that, in fact,  $D_j$ should satisfy the axioms of S5,
since its semantics is determined by $\sim_j$, which is an equivalence
relation.  This is not quite true.
The problem is with the so-called {\em truth axiom\/} of S5, which, in
this context, would say that anything that is definitely true according
to agent $j$ is true.  This would be true if there were only one agent,
but is not true with many agents, because of the different $\pi_i$
operators.  

To see the problem, suppose that $p$ is a primitive proposition.
It is easy to see that $(M,w,i) \sat D_i p \rimp p$ for all worlds $w$.
However, it is not necessarily the case that $(M,w,i) \sat D_j p \rimp
p$ if $i \ne j$.
Just because, according to agent $i$, $p$ is definitely true according
to agent $j$, it does not follow that $p$ is true {\em according to
agent $i$}.  What is true in general is that $D_j \phi \rimp \phi$ is
valid for {\em agent-independent\/} formulas.  Unfortunately,
agent independence is a semantic property.  To capture this
observation as an axiom, we need a syntactic condition sufficient to
ensure that a formula is necessarily agent independent.  I observed
earlier that formulas of the form $R_j \phi$ and $D_j \phi$ are
agent-independent.  It is immediate that Boolean combination of such 
formulas are also agent-independent.  Say that a formula is {\em necessarily
agent-independent\/} if it is a Boolean combination of formulas of the
form $R_j \phi$ and $D_{j'} \phi'$ (where the agents in the subscripts
may be the same or different).  Thus, for example, $(\neg R_1 D_2 p
\land D_1 p) \lor R_2 p$ is necessarily agent-independent.  Clearly,
whether a formula is necessarily agent-independent depends only on the
syntactic form of the formula.  Moreover, $D_j \phi \rimp \phi$ is valid
for formulas that are necessarily agent-independent.  However, 
this axiom does not capture the fact that $(M,w,i) \sat
D_i \phi \rimp \phi$ for all worlds $w$.  Indeed, this fact is not directly
expressible in the logic, but something somewhat similar is.  For arbitrary
formulas $\phi_1, \ldots, \phi_n$, note that at least one of $D_i \phi_1
\rimp \phi_1$, \ldots, $D_n \phi_n \rimp \phi_n$ must be true
respect to each triple $(M,w,i)$, $i = 1, \ldots, n$.  Thus, the formula
$(D_1 \phi_1 \rimp \phi_1) \lor \ldots \lor (D_n \phi_n \rimp \phi_n)$
is valid.  This additional property turns out to be exactly what is
needed to provide a complete axiomatization.

Let AX be the axiom system that consists of the following axioms Taut,
R1--R4, and D1--D6, and rules of inference Nec$_{{\rm R}}$, Nec$_{{\rm
D}}$, and MP: 
\begin{description}
\item[Taut.] All instances of propositional tautologies.
\item[R1.] $R_j(\phi \rimp \psi) \rimp (R_j\phi \rimp R_j\psi)$.
\item[R2.] $R_j\phi \rimp R_j R_j \phi$.
\item[R3.] $\neg R_j \phi \rimp R_j \neg R_j \phi$.
\item[R4.] $\neg R_j (\false)$.
\item[D1.] $D_j(\phi \rimp \psi) \rimp (D_j\phi \rimp D_j\psi)$.
\item[D2.] $D_j\phi \rimp D_j D_j \phi$.
\item[D3.] $\neg D_j \phi \rimp D_j \neg D_j \phi$.
\item[D4.] $\neg D_j (\false)$.
\item[D5.] $D_j \phi \rimp \phi$ if $\phi$ is necessarily
agent-independent.
\item[D6.] $(D_1 \phi_1 \rimp \phi_1) \lor \ldots \lor (D_n \phi_n \rimp
\phi_n)$. 
\item[Nec$_{{\rm R}}$.] From $\phi$ infer $R_j \phi$.
\item[Nec$_{{\rm D}}$.] From $\phi$ infer $D_j \phi$.
\item[MP.] From $\phi$ and $\phi \rimp \psi$ infer $\psi$.
\end{description}
Using standard techniques of modal
logic, it is can be shown that AX characterizes ${\cal
L}^{{\it DR}}_n$.

\thm\label{completeax} AX is a sound and complete axiomatization with
respect to vagueness structures for the language ${\cal L}^{{\it DR}}_n$.
\ethm

This shows that the semantics that I have given implicitly
assumes that agents have perfect introspection and are logically
omniscient. Introspection and logical
omniscience are both strong requirements.  There are standard techniques
in modal logic that make it possible to give semantics to $R_j$ that is 
appropriate for non-introspective agents.  With more effort, it is also
possible to avoid logical omniscience. (See, for
example, the discussion of logical omniscience in \cite{FHMV}.)
In any case,  very little of my treatment of vagueness depends on these
properties of $R_j$.

The complexity of the validity and satisfiability problem for the 
${\cal L}^{{\it DR}}_n$ can also be determined using standard
techniques.
\thm For all $n \ge 1$, determining the problem of determining the
validity (or satisfiability) of formulas in ${\cal L}^{{\it DR}}_n$ is
PSPACE-complete. 
\ethm
\prf The validity and satisfiability problems for KD45 and S5 in the
case of two or more agents is known to be PSPACE-complete \cite{HM2}.
The modal operators $R_j$ and $D_j$ act essentially like KD45 and S5
operators, respectively.  Thus, even if there is only one agent, there
are two modal operators, and a
straightforward modification of the lower bound argument in \cite{HM2}
gives the PSPACE lower bound.  The techniques of \cite{HM2} also give
the upper bound, for any number of agents. \eprf

\subsection{Capturing Vagueness and the Sorites Paradox}
Although I have described this logic as one for capturing features of
vagueness, the question still remains as to what it means to say that a
proposition $\phi$ is vague.  I suggested earlier that a standard view has
been to take $\phi$ to be vague if, in some situations, some agents 
report $\phi$ while others report $\neg \phi$, or if the same agent may
sometimes report $\phi$ and sometimes report $\neg \phi$ in the same
situation.  Both intuitions can be captured in the logic.  It is
perfectly consistent that 
$(M,w) \sat R_i \phi \land R_j \neg \phi$ if $i \ne j$; that is, the
logic makes it easy to express that two agents may 
report different things regarding $\phi$.  
Note that this depends critically on the fact that what agent $i$
reports depends 
only on the worlds in $P_i$ that agent $i$ considers plausible.  
$R_i \phi \land R_j \neg \phi$ holds only at a world $w$ where $P_i \inter
\R_i(w) \inter P_j \inter \R_j(w) = \emptyset$.  This cannot happen if
$P_i = P_j = W$, since then $w \in \R_i(w) \inter \R_j(w)$.
Expressing the second intuition
requires a little more care; it is certainly not consistent to have
$(M,w) \sat R_j \phi \land \neg R_j \phi$.  However, a more reasonable
interpretation of the second intuition is to say that in the same {\em
objective\/} situation, an agent $i$ may both report $\phi$ and $\neg \phi$.
It is certainly consistent that there are two worlds $w$ and $w'$ such
As a consequence, $(M,w) \sat \neg D_j R_j \phi$.  This statement just
says that the objective world does not determine an agent's report.  
In particular, a formula such as $\phi \land \neg D_j R_j \phi$ is
consistent; if $\phi$ is true then an agent will not necessarily report
it as true.  This can be viewed as one of the hallmarks of vagueness.
I return to this point in Section~\ref{sec:Williamson}.  

Having borderline cases has often taken to be a defining characteristic
of vague predicates.  Since I am considering a two-valued logic,
propositions do not have borderline cases: at every world, either $\phi$
is true or it is false.  However, it is not the case that $\phi$ is
either {\em definitely\/} true or false.  That is, there are borderline
cases between $D\phi$ and $D \neg \phi$.

Although the logic and the associated semantics can capture
features of vagueness, the 
question still remains as to whether it gives any insight into the problems
associated with vagueness.  I defer the discussion of some of the
problems (e.g., higher-order vagueness) to Section~\ref{sec:other}.
Here I show how it can deal with the sorites paradox.  Before going into
details, it seems to me that there should be two components to a
solution to the sorites paradox.  The first is to show where the
reasoning that leads to the paradox goes wrong in whatever formalism is
being used.  The second is to explain why, nevertheless, the argument
seems so reasonable and natural to most people.

The sorites
paradox is typically formalized as follows:
\begin{enumerate}
\item {\bf Heap}(1,000,000).
\item $\forall n > 1(\textbf{Heap}(n) \rimp \textbf{Heap}(n-1))$.
\item $\neg \textbf{Heap}(1)$.
\end{enumerate}
It is hard to argue with statements 1 and 3, so the obvious place to
look for a problem is in statement 2, the inductive step.  And, indeed,
most authors have, for various reasons, rejected this step (see, for
example, \cite{Dummett75,Sorenson01,Williamson94} for typical
discussions).  As I suggested in the introduction, it appears that
rejecting the inductive step requires committing to the existence of an
$n$ such that $n$ grains of sand is a heap and $n-1$ is not.   
While I too reject the inductive step, it does {\em not\/} follow that
there is such an $n$ in the framework I have introduced here,
because I do not assume an objective notion of heap (whose extension is
the set of natural numbers $n$ such that $n$ grains of sands form a
heap).  What constitutes a heap in my framework depends not only on the
objective aspects of the world (i.e., the number of grains of sand), but
also on the agent and her subjective state.  

To be somewhat more formal, assume for simplicity that there is only one
agent.  Consider models where the objective part of the world includes
the number of grains of sand in a particular pile of sand being observed
by the agent, and the agent's subjective state includes how many times
the agent has been asked whether a particular pile of sand constitutes a
heap.  What I have in mind here is that the sand is repeatedly added to
or removed from the pile, and each time this is done, the agent is asked
``Is this a heap?''.   Of course, the objective part of the world may
also include the shape of the pile and the lighting conditions, while
the agent's subjective state may include things like the agent's sense
perception of the pile under some suitable representation.  Exactly what
is included in the objective and subjective parts of the world do
not matter for  this analysis.

In this setup, rather than being interested in whether a pile of $n$ grains of
sand constitutes a heap, we are interested in the question of whether,
when viewing a pile of $n$ grains of sand, the agent would report that
it is a heap.  That is, we are interested in the formula $S(n)$, which I
take to be an abbreviation of $\textbf{Pile}(n) \rimp R(\textbf{Heap})$.
The formula  $\textbf{Pile}(n)$ is true at a world $w$ if, according to the
objective component of $w$, there are in fact $n$ grains of sand in the
pile.  Note that $\textbf{Pile}$ is not a vague predicate at all, but an 
objective statement about the number of grains of sand present.%
\footnote{While I
am not assuming that the agent knows the number of grains of sand
present, it would actually not affect my analysis at all if the agent
was told the exact number.}  By way of contrast, \textbf{Heap} is vague;
its truth depends on both the objective 
situation (how many grains of sand there actually are) and the agent's
subjective state.

There is no harm in restricting to models where $S(1,000,000)$ holds in all
worlds and $S(1)$ is false in all worlds where the pile actually does
consist of one grain of sand.   
If there are actually 1,000,000 grains of sand in the pile,
then the agent's subjective state is surely such that she would report
that there is a heap; and if there is actually only one grain of sand,
then the agent would surely report that there is not a heap. 
We would get the paradox if the inductive step, $\forall n >1 (S(n)
\rimp S(n-1))$ holds in all worlds.  However, it does not, for reasons
that have nothing to do with vagueness.  Note that in each world, ${\bf
Pile}(n)$ holds for exactly one value of $n$.   Consider a world $w$
where there is 1 grain 
of sand in the pile and take $n=2$.  Then $S(2)$ holds vacuously
(because its antecedent $\textbf{Pile}(2)$ is false), while $S(1)$ is
false, since in a world with 1 grain of sand, by assumption, the agent
reports that there is not a heap.  

The problem here is that the inductive statement 
$\forall n > 1 (S(n)
\rimp S(n-1))$ does not correctly capture the intended inductive
argument.  Really what we mean is more like ``if there are $n$ grains of
sand and the agent reports a heap, then when one grain of sand is
removed, the agent will still report a heap''.  

Note that removing a
grain of sand changes both the objective and subjective components of
the world.  It changes the objective component because there is one less
grain of sand; it changes the subjective component even if the agent's
sense impression of the pile remains the same, because the agent has
been asked one more question regarding piles of sand.  The change in the
agent's subjective state may not be uniquely determined, since the
agent's perception of a pile of $n-1$ grains of sand is not necessarily
always the same.  But even if it is uniquely determined, the rest of my
analysis holds.  In any case, given that the world changes, a reasonable
reinterpretation of the inductive statement might be ``For all worlds
$w$, if there are $n$ 
grains of sand in the pile in $w$, and the agent reports that there is a
heap in $w$, then the agent would report that there is a heap in all the 
worlds that may result after removing one grain of sand.''
This reinterpretation of the inductive hypothesis cannot be expressed in
the logic, but the logic could easily be extended with
dynamic-logic like operators so as to be able to express it, using a
formula such as $$\textbf{Pile}(n) \land R(\textbf{Heap}) \rimp [\mbox{remove
1 grain}](\textbf{Pile}(n-1) \land R(\textbf{Heap}).$$
Indeed, with this way of expressing the inductive step, there is no need
to include $\textbf{Pile}(n)$ or $\textbf{Pile}(n-1)$ in the formula;
it suffices to write $R(\textbf{Heap}) \rimp [\mbox{remove
1 grain}]R(\textbf{Heap})$.

Is this revised inductive step valid?  Again, it is not hard to see it
is not. Consider a world where there is a pile of 1,000,000
grains of sand, and the agent is asked for the first time whether this
is a heap.  By assumption, the agent reports that it is.  As more and
more grains of sand are removed, at some point the agent (assuming that
she has the patience to stick around for all the questions)
is bound to say that it is no longer a heap.%
\footnote{There may well be
an in-between period where the agent is uncomfortable about having to
decide whether the pile is a heap.  As I observed earlier, the semantics
implicitly assumes that the agent is willing to answer all questions
with a ``Yes'' or ``No'', but it is easy to modify things so as to allow
``I'm not prepared to say''.  The problem of vagueness still remains: At
what point does the agent first start to say ``I'm not prepared to say''?}

Although the framework makes it clear that the induction fails (as it
does in other approaches to formulating the problem), the question still
remains as to why people accept the inductive argument so quickly.  One
possible answer 
may be that it is ``almost always'' true, although making this precise
would require having a probability measure or some other measure of
uncertainty on possible worlds.  While this may be part of the answer, I
think there is another, more natural explanation.  When people are asked
the question, they do not consider worlds where they have been asked the
question many times before.  They are more likely to interpret it as
``If, in a world where I have before never been asked
whether there is a heap, I report that there is a heap, then I will
still report that there is a heap after one grain of sand is removed.''
Observe that this interpretation of the induction hypothesis is
consistent with the agent always reporting that a pile of  1,000,000
grains of sand is a heap and always reporting that a pile of 1 grain is
not a heap.  More importantly, I suspect that the inductive hypothesis
is in fact empirically true.  After an agent has
invested the ``psychic energy'' to determine whether there is a heap for
the first time, it seems to me quite likely that she will not change her
mind after one grain of sand is removed.  While this will not continue
to be true as more and more grains of sand are removed, it does not seem
to me that this is not what people think of when they answer the
question.

Graff \citeyear{Graff00} points out that a solution to the sorites
paradox that denies the truth of the inductive step, then it must deal
with three problems:
\begin{itemize}
\item The semantic question: If the inductive step is not true, is its
negation true?  If so, then is there a sharp boundary where the
inductive step fails? If
not, then what revision of classical logic must be made to accommodate
this fact?
\item The epistemological question: If the inductive step is not true,
why are we unable to say which one of its instances is not true?
\item The psychological question: If the inductive step is not true,
then why are we so inclined to accept it?
\end{itemize}
I claim that the solution I have presented here handles all these
problems well.  As we have seen, the semantic question is dealt with
since there is no sharp boundary, in the sense that (in the case of the
sorites paradox) there is an $n$ such that an agent will always say that
$n$ is a heap and $n-1$ is not (although the underlying logic presented
here is two-valued).  As for the epistemological question, we cannot say
which instances of the inductive step are true because the truth of a
particular step depends on the agent's subjective state.  
A particular instance might be true in one subjective state and not another.
Moreover, as I pointed out in the discussion of intransitivity, the
response may not even be a deterministic function of the agent's state.
This makes it impossible to say exactly how an agent will respond to a
particular sorites sequence.  Finally, I have given an argument as to
why we are inclined to accept the truth of the inductive step, which
depends on a claim of how the statement is interpreted.

\section{Relations to Other Approaches}\label{sec:other}

In this section I consider how the approach to vagueness sketched in the
previous section is related to other approaches to
vagueness that have been discussed in the literature.

\subsection{Context-Dependent Approaches}
My approach for dealing with the sorites paradox is perhaps closest to
what Graff \citeyear{Graff00} has called {\em context-dependent\/}
approaches, where the truth of a vague predicate depends on context.
The ``context'' in my approach can be viewed as a combination of the
objective state and the agent's subjective state.  Although a number of
papers have been written on this approach (see, for example,
\cite{Graff00,Kamp75,Soames99}), perhaps the closest in spirit to mine is
that of Raffman \citeyear{Raffman94}.

In discussing sorites-like paradoxes, Raffmman considers a sequence of
colors going gradually from red to orange, and assumes that to deal with
questions like ``if patch $n$ is red, then so is patch $n-1$'', the
agent makes pairwise judgments.  She 
observes that it seems reasonable that an agent will always place patches
$n$ and $n+1$, judged at the same time, in  same category (both red,
say, or both orange).  However, it is plausible that patch $n$ will be
assigned different colors when paired with $n-1$ than when paired with
$n+1$.   This observation (which I agree is likely to be
true) is easily accommodated in the framework that I have presented
here: If the agent's subjective state includes the perception of two
adjacent color patches, and she is asked to assign both a color, then
she will almost surely assign both the same color.  
Raffman also observes that the color judgment may depend on the
colors that have already been seen as well as other random features (for
example, how tired/bored the agent is), although she does not
consider the specific approach to the sorites paradox that I do (i.e.,
the interpretation of the inductive step of the paradox as ``if, {\em
the first time I am asked}, I report that $P(n)$ holds, then I will also
report that $P(n-1)$ holds if asked immediately afterwards'').

However, none of the context-dependent approaches use a model that
explicitly distinguishes the objective features of the world from the
subjective features of a world.  Thus, they cannot deal with the interplay
of the ``definitely'' and ``reports that'' operators, which plays a
significant role in my approach.  By and large, they also seem to ignore
issues of higher-order vagueness, which are well dealt with by this
interplay (see Section~\ref{sec:higher-order}).

\subsection{Fuzzy Logic}\label{sec:fuzzy}
{\em Fuzzy logic\/} \cite{Zadlog} seems like a natural approach to
dealing with vagueness, since it does not require a predicate be
necessarily true or false; rather, it can be true to a certain degree.
As I suggested earlier, this does not immediately resolve the problem
of vagueness, since a statement like ``this cup of coffee is sweet to
degree .8'' is itself a crisp statement, when the intuition suggests it
should also be vague.

Although I have based my approach on a two-valued logic, 
there is a rather natural connection between my approach and fuzzy
logic. 
We can take the degree of truth of a formula $\phi$ in world $w$
to be the fraction of agents $i$ such that $(M,w,i) \sat \phi$. 
We expect that, in most worlds, the degree of truth of a formula will be
close to either 0 or 1.  We can have meaningful communication
precisely because there is a large degree of agreement in how agents
interpret subjective notions thinness, tallness, sweetness.

Note that the degree of truth of $\phi$ in $(o,s_1,
\ldots, s_n)$ does not depend just on $o$, since $s_1, \ldots, s_n$ are
not deterministic functions of $o$.  But if we assume that each
objective situation $o$  
determines a probability distribution on tuples $(s_1, \ldots, s_n)$,
then if $n$ is large, for many predicates of interest (e.g., {\bf Thin},
{\bf Sweet}, {\bf Tall}), I expect that, as an empirical matter, the
distribution will be normally distributed with a very small variance.
In this case, the degree of truth of such a predicate in an objective
situation $o$ can be taken to be the expected degree of truth of $P$,
taken over all worlds $(o, s_1, \ldots, s_n)$ whose first component is
$o$.   

This discussion shows that my approach to vagueness is compatible with 
assigning a degree of truth in the interval $[0,1]$ to vague
propositions, as is done in fuzzy logic.  Moreover non-vague
propositions (called {\em crisp\/} in the fuzzy logic literature) get
degree of truth either 0 or 1.  However, while this is a way 
of giving a natural interpretation to degrees of truth, and it supports
the degree of truth of $\neg \phi$ being 1 minus the degree of truth of
$\phi$, as is done in fuzzy logic, it does not support the 
semantics for $\land$ typically taken in fuzzy logic, where the degree
of truth of $\phi \land \psi$ is taken to be the minimum of the degree
of truth of $\phi$ and the degree of truth of $\psi$.  Indeed, under my
interpretation of degree of truth, there is no functional connection 
between the degree of truth of $\phi$, $\psi$, and
$\phi\land \psi$

\subsection{Supervaluations}
The $D$ operator also has close relations to the notion of {\em
supervaluations} \cite{Fine75,vF68}.  Roughly speaking, the intuition
behind supervaluations is that language is not completely precise.
There 
are various ways of ``extending'' a world to make it precise.  A formula
is then taken to be true at a world $w$ under this approach if it is
true under all ways of extending the world.  Both the $R_j$ and $D_i$
operators have some of the flavor of supervaluations.  If we consider
just the objective component of a world $o$, there are various ways of
extending it with subjective components $(s_1, \ldots, s_n)$.  $D_i
\phi$ is true at an objective world $o$ if $(M,w,i) \sat \phi$ for all
worlds $w$ that extend $o$.  (Note that 
the truth of $D_j \phi$ depends only on the objective component
of a world.)  Similarly, given just a subjective
component $s_j$ of a world, $R_j \phi$ is true of $s_j$ if $(M,w,i) \sat
\phi$ for all worlds that extend $s_i$.  Not surprisingly, properties
of supervaluations can be expressed using $R_j$ or $D_j$.
Bennett \citeyear{Bennett98} has defined a modal logic that 
formalizes the supervaluation approach.

\subsection{Higher-Order Vagueness}\label{sec:higher-order}
In many approaches towards vagueness, there has been discussion of
{\em higher-order vagueness\/} (see, for example,
\cite{Fine75,Williamson94}).   In the context of the supervaluation
approach, we can say that $D \phi$ (``definitely $\phi$'') holds at a world $w$
if $\phi$ is true in all extensions of  $w$.  Then $D\phi$ is not
vague; at each world, either $D\phi$ or $\neg D \phi$ (and $D \neg D
\phi$) is true (in the
supervaluation sense).  But using this
semantics for definitely, it seems that there is a problem.  For under
this semantics, ``definitely $\phi$'' implies ``definitely definitely
$\phi$'' (for essentially the same reasons that $D_i \phi \rimp D_i D_i
\phi$ in the semantics that I have given).  But, goes the argument, this
does not allow the statement ``This is definitely red'' to be vague.  
A rather awkward approach is taken to dealing with this by Fine
\citeyear{Fine75} (see also \cite{Williamson94}), which  allows different
levels of interpretation.  

I claim that the real  problem is that higher-order vagueness should not
be represented using the modal operator $D$ in isolation.  Rather, a
combination of $D$ and $R$ should be used.  It is not interesting
particularly to ask when it is definitely the case that 
it is definitely the case that something is red.  This is indeed true
exactly if it is definitely red.  What is more interesting is 
when it is definitely the case that agent $i$ would report that
an object is definitely red.  This is represented by the formula $D_i
R_i D_i \textbf{Red}$.  We can iterate and ask when $i$ would report
that it is definitely the case that he would report that it is
definitely the case that he would report it is definitely red,
i.e., when $D_i R_i D_i R_i D_i \textbf{Red}$ holds, and so on.  
It is easy to see that 
$D_i R_i p$ does not imply $D_i R_i D_i R_i p$; lower-order vagueness
does not imply higher-order vagueness.  Since I have assumed that agents
are introspective, it can be shown that higher-order vagueness implies
lower-order vagueness.  In particular, 
$D_i R_i D_i R_i \phi$
does imply $D_i R_i \phi$. (This follows using the fact that $D_i \phi
\rimp \phi$ and $R_i R_i \phi \rimp R_i \phi$ are both valid.)  
The bottom line here is that by separating the $R$ and $D$ operators in
this way, issues of higher-order vagueness become far less vague.

\subsection{Williamson's Approach}\label{sec:Williamson}
One of the leading approaches to vagueness in the recent literature is
that of Williamson; see \cite[Chapters 7 and 8]{Williamson94} for an
introduction.  Williamson considers an
epistemic approach, viewing vagueness as ignorance.  Very roughly
speaking, he uses ``know'' where I use ``report''.  However, he insists
that it cannot be the case that if you know something, then you know you know
it, whereas my notion of reporting has the property that $R_i$ implies
$R_i R_i$.  It is instructive to examine the example that Williamson
uses to argue that you cannot know what you know, to see where his
argument breaks down in the framework I have presented.

Williamson considers a situation where you look at a crowd and do not
know the number of people in it.  He makes what seem to be a number of
reasonable assumptions.  Among them is the following:
\begin{quote}
I know that if there are exactly $n$ people, then I do not know that
there are not exactly $n-1$ people.
\end{quote}
This may not hold in my framework.  This is perhaps easier to see if we
think of a robot with sensors.  If there are $n$ grains of sugar in the
cup, it is possible that a sensor reading compatible with $n$ grains will
preclude there being $n-1$ grains.  For example, suppose that, as in
Section~\ref{sec:intransitivity}, if there are $n$ grains of sugar, and
the robot's sensor reading is between $\lfloor (n-4)/10 \rfloor$ and
$\lfloor 
(n+4)/10 \rfloor$.  If there are in fact 16 grains of sugar, then the
sensor reading could be 2 ($= \lfloor (16 + 4)/10 \rfloor$).  But if the
robot knows how its sensor works, then if its
sensor reading is 2, then it knows that if there are
exactly 16 grains of sand, then (it knows that) 
there are not exactly 15 grains of sugar.  Of course, it is
possible to change the semantics of $R_i$ so as to validate Williamson's
assumptions.  But this point seems to be orthogonal to dealing with
vagueness. 

Quite apart from his treatment of epistemic matters, Williamson
seems to implicitly assume that there is an objective notion of what I
have been calling subjectively vague notions, such as red, sweet, and
thin.  This
is captured by what he calls the {\em supervenience thesis},
which roughly says that if two worlds agree
on their objective part, then they must agree on how they interpret what
I have called subjective  propositions.
Williamson focuses on the example of thinness, in which case
his notion of supervenience implies that
``If $x$ has exactly the same physical measurements in a possible
situation $s$ and $y$ has in a possible situation $t$, then $x$ is thin
in $s$ if and only if $y$ is thin in $t$'' \cite[p. 203]{Williamson94}. 
I have rejected this viewpoint here, since, for me, whether $x$ is this
depends also on the agent's subjective state.
Indeed, rejecting this viewpoint is a
central component of my approach to intransitivity and vagueness.

Despite these differences, there is one significant point of contact
between Williamson's approach and that presented here.  Williamson
suggests modeling vagueness using a modal operator $C$ for {\em
clarity}.  Formally, he takes a model $M$ to be a quadruple
$(W,d,\alpha,\pi)$, where $W$ is a set of worlds and $\pi$ is an
interpretation as above (Williamson seems to implicitly assume that
there is a single agent), where $d$ is a metric on $W$ (so that $d$
is a symmetric function mapping
$W \times W$ to $[0,\infty)$ such that $d(w,w') = 0$ iff $w = w'$
and $d(w_1,w_2) + d(w_2,w_3) \le d(w_1,w_3)$), and $\alpha$ is a
non-negative real number.  The semantics of formulas is defined in the
usual way; the one interesting clause is that for $C$:
$$\begin{array}{l}
(M,w) \sat C\phi \mbox{ iff }
\mbox{\ \ \ } (M,w')\sat \phi \mbox{ for all $w'$ such
that $d(w,w') \le \alpha$}.
\end{array}$$
Thus, $C\phi$ is true at a world $w$ if $\phi$ is true at all worlds
within $\alpha$ of $w$.

The intuition for this model is perhaps best illustrated by considering
it in the framework discussed in the previous section, assuming
that there is only one proposition, say  \textbf{Tall(TW)}, and one agent.
Suppose that {\bf Tall(TW)} is taken to hold if TW is above some
threshold height $t^*$.  Since {\bf Tall(TW)} is the only primitive
proposition, we can take the objective part of a world to be determined
by the actual height of TW.  For simplicity, assume that the agent's
subjective state is determined by the agent's subjective estimate of
TW's height (perhaps as a result of a measurement).  Thus, a world can
be taken to be a tuple $(t,t')$, where $t$ is TW's height and $t'$ is
the agent's subjective estimate of the height.  Suppose that the agent's
estimate is within $\alpha/2$ of TW's actual height,
so that the set $W$ of possible worlds consists of all  pairs $(t,t')$
such that $|t-t'| \le \alpha/2$.  
Assume that all worlds are plausible (so that
$P = W$).  It is
then easy to check that $(M,(t,t')) \sat DR({\bf Tall(TW)})$ iff $t \ge 
t^* + \alpha$.  That is, the agent will definitely say that TW is Tall
iff TW's true height is at least $\alpha$ more than the threshold $t^*$
for tallness, since in such worlds, the agent's subjective estimate of
TW's height is
guaranteed to be at least $t^* + \alpha/2$.  

To connect this to Williamson's model, suppose that the metric $d$ is
such that $d((t,t'),(u,u')) = |t-u|$; that is, the distance between
worlds is taken to be the difference between TW's actual height in these
worlds.  Then it is immediate that $(M,(t,t')) \sat C(\textbf{Tall(TW)})$ iff
$t \ge t^* + \alpha$.  
In fact, a more general statement is true.  By definition,
$(M,(t,t')) \sat C \phi$ iff $(M,(u,u')) \sat \phi$ for all $(u,u') \in
W$ such that $|t-u| \le \alpha$.  It is easy to check that 
$(M,(t,t')) \sat DR \phi$ iff $(M,(u,u')) \sat \phi$ for all $(u,u') \in
W$ such that $|t - u'| \le \alpha/2$.  Finally, a straightforward
calculcation shows that, for a fixed $t$, 
$$\{u: \exists u' ((u,u') \in W, |t-u| \le \alpha)\} = 
\{u: \exists u' ((u,u') \in W, |t-u'| \le \alpha/2)\}.$$
Thus, if $\phi$ is a formula whose truth depends just on the objective
part of the world (as is the case for {\bf Tall(TW)} as I have defined
it) then $(M,(t,t')) \sat C\phi$ iff $(M,(t,t') \sat DR \phi$.

\commentout{
Williamson does not give examples of how the metric should be
interpreted.  Under my interpretation, it would be more reasonable to
have a different metric for each proposition.  But ignoring this point,
the agreement between $C$ is $DR$ is more than just a superficial one.
For example, Williamson provides a complete axiomatization of $C$; it is
not hard to show that if all worlds are plausible (that is, if $P = W$,
so that $R$ becomes an S5 relation), then the combination $DR$ is
characterized by exactly the same axioms as $C$.%
\footnote{Proof sketch: Williamson shows that $C$ is characterized by
the logic KTB.  Thus, we can view $C$ as being defined by a binary
relation that is reflexive and symmetric.  Given the assumptions here,
$D$ and $R$ are separately characterized by the logic S5, and thus are
represented by equivalence relations.  The composition of two
equivalence relations is easily seen to be reflexive and symmetric}
}

Williamson suggests that a proposition $\phi$ should be taken to be vague
if $\phi \land \neg C\phi$ is satisfiable.  In Section~\ref{sec:vagueness}, I
suggested that $\phi \land \neg DR \phi$ could be taken as one of the
hallmarks of vagueness.  Thus, I can capture much the same
intuition for vagueness as Williamson by using $DR$ instead of $C$,
without having to make what seem to me unwarranted epistemic assumptions.

\section{Discussion}\label{sec:discussion}

I have introduced what seems to me a natural approach to dealing with
intransitivity of preference and vagueness.  Although various pieces of
the approach seem certainly have appeared elsewhere, it seems that this
particular packaging of the pieces is novel.  The approach leads to a
straightforward logic of vagueness, while avoiding many of the problems
that have plagued other approaches.  In particular, it gives what I
would argue is a clean solution to the semantic, epistemic, and
psychological problems associated with vagueness, while being able to
deal with higher-order vagueness.

\commentout{
One thing that I have not discussed is why we accept arguments like the
sorites paradox seem so plausible.  That is, even knowing that notions
like ``heap'' are vague, so that there will be disagreement over where
the crossover between ``heap'' and ``not heap'' will be, the claim that
if $n+1$ grains of sand make a heap then so do $n$ grains seems quite
plausible.  There is an intuition that this statement is ``almost always'',
which I believe needs to be distinguished from vagueness.  One way of
making this intuition precise is using probability.  Suppose that we put
a probability distribution on worlds.  It seems reasonable to expect
that with high probability, this statement is true, but that might be
the case for trivial reasons, even if there is a sharp cutoff between
heapness and non-heapness.  To understand the issue here, consider, by
way of contrast, the statement that of a cup of water that if it is
liquid at $n$ degrees Celsius, then it is still liquid at $n-1$ degrees.
Although in fact the cutoff between water and ice is not completely
sharp (at $0\deg$ Celsius, there will typically be some combination of
water and ice in the cup), the transition happens in a very narrow
range.  In the case of heaps, it is likely to be true that for all
objective situations, the probability 
}

\paragraph{Acknowledgments:} I'd like to thank 
Delia Graff, Rohit Parikh, Riccardo
Pucella, and Tim Williamson for comments on a previous draft of the paper.

\bibliographystyle{chicago}
\bibliography{z,joe}
\end{document}

%% file: defn.tex
\newtheorem{THEOREM}{Theorem}[section]
\newenvironment{theorem}{\begin{THEOREM} \hspace{-.85em} {\bf :} }%
                        {\end{THEOREM}}
\newtheorem{LEMMA}[THEOREM]{Lemma}
\newenvironment{lemma}{\begin{LEMMA} \hspace{-.85em} {\bf :} }%
                      {\end{LEMMA}}
\newtheorem{COROLLARY}[THEOREM]{Corollary}
\newenvironment{corollary}{\begin{COROLLARY} \hspace{-.85em} {\bf :} }%
                          {\end{COROLLARY}}
\newtheorem{PROPOSITION}[THEOREM]{Proposition}
\newenvironment{proposition}{\begin{PROPOSITION} \hspace{-.85em} {\bf :} }%
                            {\end{PROPOSITION}}
\newtheorem{DEFINITION}[THEOREM]{Definition}
\newenvironment{definition}{\begin{DEFINITION} \hspace{-.85em} {\bf :} \rm}%
                            {\end{DEFINITION}}
\newtheorem{CLAIM}[THEOREM]{Claim}
\newenvironment{claim}{\begin{CLAIM} \hspace{-.85em} {\bf :} \rm}%
                            {\end{CLAIM}}
\newtheorem{EXAMPLE}[THEOREM]{Example}
\newenvironment{example}{\begin{EXAMPLE} \hspace{-.85em} {\bf :} \rm}%
                            {\end{EXAMPLE}}
\newtheorem{REMARK}[THEOREM]{Remark}
\newenvironment{remark}{\begin{REMARK} \hspace{-.85em} {\bf :} \rm}%
                            {\end{REMARK}}
\newcommand{\thm}{\begin{theorem}}
\newcommand{\lem}{\begin{lemma}}
\newcommand{\pro}{\begin{proposition}}
\newcommand{\dfn}{\begin{definition}}
\newcommand{\rem}{\begin{remark}}
\newcommand{\xam}{\begin{example}}
\newcommand{\cor}{\begin{corollary}}
\newcommand{\prf}{\noindent{\bf Proof:} }
\newcommand{\ethm}{\end{theorem}}
\newcommand{\elem}{\end{lemma}}
\newcommand{\epro}{\end{proposition}}
\newcommand{\edfn}{\bbox\end{definition}}
\newcommand{\erem}{\bbox\end{remark}}
\newcommand{\exam}{\bbox\end{example}}
\newcommand{\ecor}{\end{corollary}}
\newcommand{\eprf}{\bbox\vspace{0.1in}}
\newcommand{\beqn}{\begin{equation}}
\newcommand{\eeqn}{\end{equation}}
\newcommand{\bbox}{\vrule height7pt width4pt depth1pt}

\newcommand{\clm}{\begin{claim}}
\newcommand{\eclm}{\end{claim}}
\newcommand{\sat}{\models}

\newcommand{\rimp}{\Rightarrow}
\newcommand{\inter}{\cap}
\renewcommand{\phi}{\varphi}
\newcommand{\R}{{\cal R}}
\newcommand{\ol}{\setlength{\itemsep}{0pt}\begin{enumerate}}
\newcommand{\eol}{\end{enumerate}\setlength{\itemsep}{-\parsep}}
\newcommand{\ul}{\setlength{\itemsep}{0pt}\begin{itemize}}
\newcommand{\dl}{\setlength{\itemsep}{0pt}\begin{description}}
\newcommand{\edl}{\end{description}\setlength{\itemsep}{-\parsep}}
\newcommand{\eul}{\end{itemize}\setlength{\itemsep}{-\parsep}}

\newcommand{\false}{\mbox{{\it false}}}

\newcommand{\commentout}[1]{}

\newcommand{\bi}{\begin{itemize}}
\newcommand{\ei}{\end{itemize}}
\newcommand{\be}{\begin{enumerate}}
\newcommand{\ee}{\end{enumerate}}

%% file: vaguecorr.bbl
\begin{thebibliography}{}

\bibitem[\protect\citeauthoryear{Aumann}{Aumann}{1976}]{Au}
Aumann, R.~J. (1976).
\newblock Agreeing to disagree.
\newblock {\em Annals of Statistics\/}~{\em 4\/}(6), 1236--1239.

\bibitem[\protect\citeauthoryear{Bennett}{Bennett}{1998}]{Bennett98}
Bennett, B. (1998).
\newblock Modal semantics for knowledge bases dealing with vague concepts.
\newblock In {\em Principles of Knowledge Representation and Reasoning:
  Proc.~Sixth International Conference (KR '98)}, pp.\  234--244.

\bibitem[\protect\citeauthoryear{Black}{Black}{1937}]{Black37}
Black, M. (1937).
\newblock Vagueness: an exercise in logical analysis.
\newblock {\em Philosophy of Science\/}~{\em 4}, 427--455.

\bibitem[\protect\citeauthoryear{Dummett}{Dummett}{1975}]{Dummett75}
Dummett, M. (1975).
\newblock Wang's paradox.
\newblock {\em Synthese\/}~{\em 30}, 301--324.

\bibitem[\protect\citeauthoryear{Fagin, Halpern, Moses, and Vardi}{Fagin
  et~al.}{1995}]{FHMV}
Fagin, R., J.~Y. Halpern, Y.~Moses, and M.~Y. Vardi (1995).
\newblock {\em Reasoning about Knowledge}.
\newblock Cambridge, Mass.: MIT Press.

\bibitem[\protect\citeauthoryear{Fine}{Fine}{1975}]{Fine75}
Fine, K. (1975).
\newblock Vagueness, truth, and logic.
\newblock {\em Synthese\/}~{\em 30}, 265--300.

\bibitem[\protect\citeauthoryear{Friedman and Halpern}{Friedman and
  Halpern}{1994}]{FrH1}
Friedman, N. and J.~Y. Halpern (1994).
\newblock A knowledge-based framework for belief change. {P}art {I}:
  foundations.
\newblock In {\em Theoretical Aspects of Reasoning about Knowledge: Proc.~Fifth
  Conference}, pp.\  44--64.

\bibitem[\protect\citeauthoryear{Graff}{Graff}{2000}]{Graff00}
Graff, D. (2000).
\newblock Shifting sands: an interest-relative theory of vagueness.
\newblock {\em Philophical Topics\/}~{\em 28\/}(1), 45--81.

\bibitem[\protect\citeauthoryear{Halpern}{Halpern}{1997}]{Hal15}
Halpern, J.~Y. (1997).
\newblock On ambiguities in the interpretation of game trees.
\newblock {\em Games and Economic Behavior\/}~{\em 20}, 66--96.

\bibitem[\protect\citeauthoryear{Halpern}{Halpern}{2001}]{Hal35}
Halpern, J.~Y. (2001).
\newblock Sleeping beauty reconsidered: {C}onditioning and reflection in
  asynchronous systems.
\newblock {\em Principles of Knowledge Representation and Reasoning:
  Proc.~Ninth International Conference (KR '04)\/}~{\em 37}, 12--22.
\newblock To appear, {\em Oxford Studies in Epistemology, vol. 1}.

\bibitem[\protect\citeauthoryear{Halpern and Fagin}{Halpern and
  Fagin}{1989}]{HFfull}
Halpern, J.~Y. and R.~Fagin (1989).
\newblock Modelling knowledge and action in distributed systems.
\newblock {\em Distributed Computing\/}~{\em 3\/}(4), 159--179.

\bibitem[\protect\citeauthoryear{Halpern and Moses}{Halpern and
  Moses}{1992}]{HM2}
Halpern, J.~Y. and Y.~Moses (1992).
\newblock A guide to completeness and complexity for modal logics of knowledge
  and belief.
\newblock {\em Artificial Intelligence\/}~{\em 54}, 319--379.

\bibitem[\protect\citeauthoryear{Hempel}{Hempel}{1939}]{Hempel39}
Hempel, C.~G. (1939).
\newblock Vagueness and logic.
\newblock {\em Philosophy of Science\/}~{\em 6}, 163--180.

\bibitem[\protect\citeauthoryear{Kamp}{Kamp}{1975}]{Kamp75}
Kamp, H. (1975).
\newblock Two theories about adjectives.
\newblock In E.~L. Keenan (Ed.), {\em Formal Semantics of Natural Language},
  pp.\  123--155. Cambridge, U.K.: Cambridge University Press.

\bibitem[\protect\citeauthoryear{Keefe}{Keefe}{2000}]{Keefe00}
Keefe, R. (2000).
\newblock {\em Theories of Vaguness}.
\newblock Cmabridge, UK: Cambridge University Press.

\bibitem[\protect\citeauthoryear{Keefe and Smith}{Keefe and Smith}{1996}]{KS96}
Keefe, R. and P.~Smith (1996).
\newblock {\em Vaguness: A Reader}.
\newblock Cambridge, Mass.: MIT Press.

\bibitem[\protect\citeauthoryear{Peirce}{Peirce}{1956}]{Peirce31}
Peirce, C.~S. (1931--1956).
\newblock {\em Collected Writings of Charles Sanders Peirce}.
\newblock Cambridge, Mass.: Harvard University Press.
\newblock Edited by C. Hartshorne, P. Weiss, et al.

\bibitem[\protect\citeauthoryear{Poincar{\'e}}{Poincar{\'e}}{1902}]{Poincare02}
Poincar{\'e}, H. (1902).
\newblock {\em La Science et l'Hypoth{\`e}se}.
\newblock Paris: Flamarrion Press.

\bibitem[\protect\citeauthoryear{Raffman}{Raffman}{1994}]{Raffman94}
Raffman, D. (1994).
\newblock Vaguenss without paradox.
\newblock {\em The Philosophical Review\/}~{\em 103\/}(1), 41--74.

\bibitem[\protect\citeauthoryear{Soames}{Soames}{1999}]{Soames99}
Soames, S. (1999).
\newblock {\em Understanding Truth}.
\newblock New York: Oxford University Press.

\bibitem[\protect\citeauthoryear{Sorenson}{Sorenson}{2001}]{Sorenson01}
Sorenson, R. (2001).
\newblock {\em Vagueness and Contradiction}.
\newblock Oxford, UK: Oxford University Press.

\bibitem[\protect\citeauthoryear{{van Fraassen}}{{van Fraassen}}{1968}]{vF68}
{van Fraassen}, B.~C. (1968).
\newblock Presuppositions, implications, and self-reference.
\newblock {\em Journal of Philosophy\/}~{\em 65}, 136--152.

\bibitem[\protect\citeauthoryear{Williamson}{Williamson}{1994}]{Williamson94}
Williamson, T. (1994).
\newblock {\em Vagueness}.
\newblock London/New York: Routledge.

\bibitem[\protect\citeauthoryear{Zadeh}{Zadeh}{1975}]{Zadlog}
Zadeh, L.~A. (1975).
\newblock Fuzzy logics and approximate reasoning.
\newblock {\em Synthese\/}~{\em 30}, 407--428.

\end{thebibliography}
